\documentclass{article}
\usepackage{spconf,amsmath,graphicx}
\usepackage{url}

\title{Visible and Infrared Image Fusion Using Encoder-Decoder Network}
\twoauthors
 {Ferhat Can ATAMAN}
	{Middle East Technical University\\
    ataman.ferhat@metu.edu.tr}
 {Gözde BOZDAĞI AKAR}
	{Middle East Technical University\\
	bozdagi@metu.edu.tr}
 
\begin{document}
%
\maketitle
\begin{abstract}
The aim of multispectral image fusion is to combine object or scene features of images with different spectral characteristics to increase the perceptual quality. In this paper, we present a novel learning-based solution to image fusion problem focusing on infrared and visible spectrum images. The proposed solution utilizes only convolution and pooling layers together with a loss function using no-reference quality metrics. The analysis is performed qualitatively and quantitatively on various datasets. The results show better performance than state-of-the-art methods. Also, the size of our network enables real-time performance on embedded devices. Project codes can be found at \url{https://github.com/ferhatcan/pyFusionSR}.  
\end{abstract}
\section{Introduction}
\label{sec:intro}

Image fusion aims to combine relative information from input images with different spatial and spectral characteristics. For example, infrared images can see through smoke whereas, visible spectrum images have a high spatial resolution. Combining the benefits of these two modalities by fusion enables us to have superior results in many applications such as object detection, recognition, and tracking.  

In the literature, there are numerous studies on image fusion of especially IR and visible images. In  \cite{survey}, the methods are divided into seven categories according to their corresponding theories, namely multi-scale transform, sparse representation, neural network, subspace, and saliency-based methods, hybrid models, and other methods. In all of these methods, source images are decomposed into several levels and corresponding layers are fused with particular rules. The fusion strategy of these methods can be complex that compares corresponding regions in the calculated features such as CNN-method or as simple as element-wise adding features such as Hybrid-MSD.
Deep neural network-based methods are one of the categories mentioned in \cite{survey}. These methods are preferred for both feature extraction, fusion, and reconstruction steps. \cite{DLF}(DLF), after decomposing images to base and details, VGG19 pre-trained network is used to extract features from both detail images and results are fused via averaging strategy. Since DLF uses a fixed neural network that is trained on the visible image spectrum to classify objects, feature extraction from IR images is the bottleneck of this algorithm.  In \cite{deepfuse}(DeepFuse), features extracted with same weights shared input networks(same neural network for both inputs) and features are fused using element-wise adding. Final CNN layers are used to reconstruct images from fused features. DeepFuse is used to solve different exposure image fusion and it uses neural networks to fuse only Y channels of YCbCr images. Weighted fusion is used for color channels, Cb and Cr. In \cite{rtfnet}(RTFNet), an end to end Encoder-Decoder network is designed. Encoder (visible and infrared separately) extracts features, these features are fused into the visible encoder and the ultimate encoded feature is connected to the decoder to reconstruct the fused image. This method is used to solve semantic segmentation problems in multi-spectral image pairs. The latest approaches Rfn-Nest \cite{rfnnest} and Dual-Branch \cite{dualbranch} also use the power of the encoder-decoder network. They directly reconstruct the fused image from source image pairs. These two use the same pre-trained encoder layer for source images which can cause loss of features to be extracted. One of the main drawbacks of such neural networks is they are computationally intensive operations and their running time is high. Furthermore, there should be enough data to train the network properly and there is also no reference fusion result. 

Another important issue in image fusion is that there is no absolute correct reference to evaluate the results. Thus, no reference image fusion quality metrics are presented to measure the accuracy of the methods applied \cite{objAssesment}. However, these evaluation metrics suffer from distinguishing results' strengths and weaknesses. In addition to objective metrics, subjective evaluations are required to examine experiments more accurately. No reference image quality metric is a different area that tries to give solutions to these issues. Deep learning methods need defining a metric called loss function to train the network in the desired way. The network should differentiate results' accuracy. In image fusion applications, we have no reference image to compare so, this makes it difficult to solve the problem with this approach. Commonly, pre-trained networks are used to extract features and classical methods are preferred to fuse these features. We try to learn extracting features of visible and infrared images, combining and reconstructing a fused image using an end-to-end deep neural network that handles all of them. 


To address both of these issues,  we present a deep convolutional neural network(DCNN)  architecture. Our main contributions are summarized below.
\begin{itemize}
    \item  We design an end-to-end trainable network. It makes architecture all-in-one so, there is no need to post or pre-process calculations. Also, in our design, we use only convolutional and pooling layers. The number of convolutional filter weights of our network design is reduced compared to the above deep learning based methods. Thus, our architecture is running fast without sacrificing performance.
    \item We used a custom loss function using no-reference quality metrics. The mean square error function is also integrated into the loss function to increase visual perception quality. In this way, generated color images become more realistic and natural. There are no absurd color changes in too dark or bright parts of the image. To measure the perceived qualities of our results, we use no reference perception quality metric mentioned in \cite{PaQ}.
\end{itemize}

\begin{figure}[ht]
	\centering
	\includegraphics[width=0.45\textwidth]{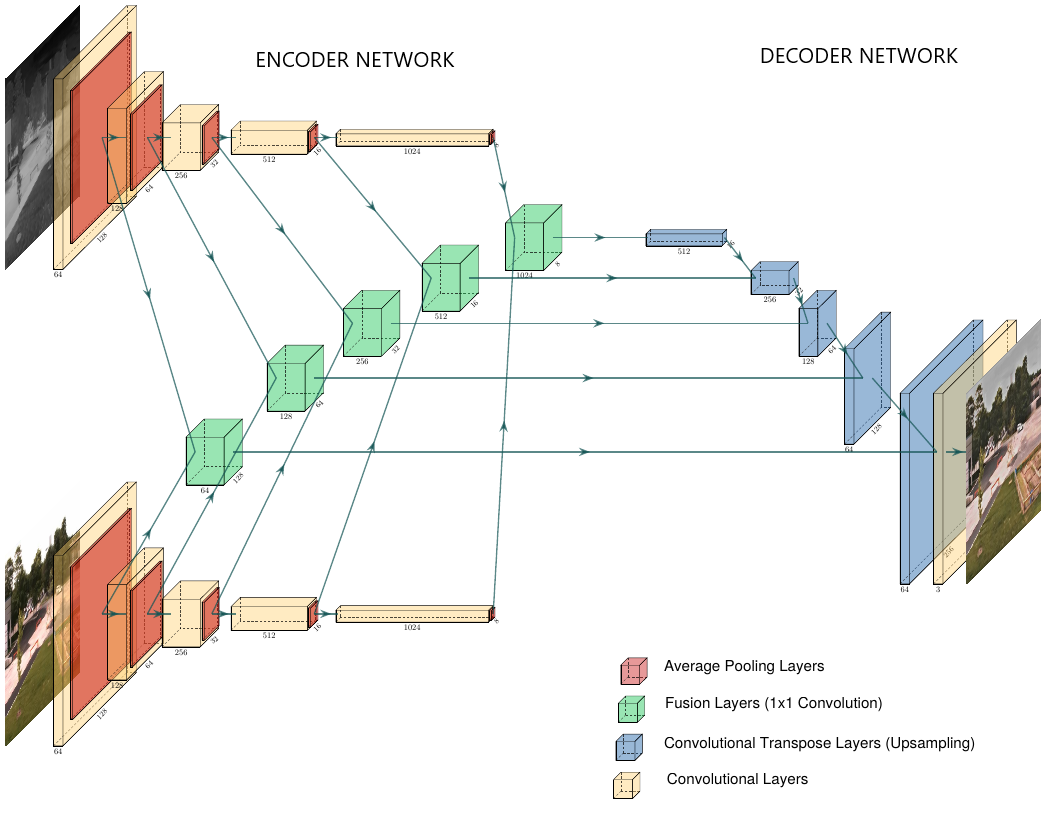}
	\caption{The proposed network architecture. The first part includes two identical encoder networks that correspond to infrared and visible images respectively. Then, extracted features are fused using 1x1 convolutional layers(green boxes in the middle). The fused features are used to construct a final image in the decoder network. }
	\label{fig:arch}
\end{figure}

\section{The Approach}

\subsection{The Architecture}

The architecture of our proposed system includes three main parts: encoder, decoder, and fusion networks. The encoder network extract features from input images. Then extracted features that are the output of each layer in the encoder network are fused using a fusion network. Then, fused outputs are given to the decoder network to generate the final result. This architecture is only composed of convolutional layers, so there is no dependency on input image shape except channel sizes. The visible image has three channels, whereas the infrared image has a single channel. Furthermore, the input images should be multiple of 32 to network works because the encoder network has 5 convolutional layers and each layer downsizes input width and height to half. Decoder Layer concatenates skip connection input and output of the previous layer that should match. Unsuitable images are resized using zero padding. The proposed network architecture is given in Fig.\ref{fig:arch}.

The encoder network downsizes the input image in each layer while increasing its channel dimensions. Each layer's output consists of compressed features of the input image. In each layer, the height and width of the input are cut in half, whereas channel size is doubled. It is achieved by average pooling after the convolutional layer. Even if each image spectrum has identical encoder network architecture, their network parameters are separate.

After features are extracted using the encoder network, these should be fused before feeding into the decoder network. This part is one of the main contributions of our design. The features of each corresponding encoder layer are concatenated through channels. That is the input channel size is doubled. Then, a convolutional layer with filter size 1 is used to fuse features and the output channel size becomes half of the input channel size. This approach is considered as an attention layer because this layer only learns channel contributions of each pixel point. If a part of the image includes information in relative pixel, the response becomes higher to include in the fused image.

Lastly, the decoder network takes fused features and generates an output image. The decoder network is the mirrored version of the encoder network. To upsample feature layers, a convolutional transpose layer is used. There is no pooling layer in the decoder. In each layer, corresponding features coming from the fusion network and output of the previous layer are concatenated before convolution operations. This is commonly known as skip connections. In the end, the hyperbolic tangent function (tanh) is used to generate pixels in the range between -1 and 1.

\subsection{Loss Functions}
There is no reference or ground truth images for the image fusion task, thus, it is crucial to choose a suitable metric that can measure the quality of the fused image by comparing input images. There exists some fusion quality metrics in the literature and our aim is to use these metrics as loss function to train our network. We use \cite{qw} $Q_w$, that compares two input images with result image in local regions.
In Eqn.\ref{Q0}, $\sigma_x$ is the square-root of variance, and $\bar{x}$ is mean of input image patch $x$. It also applies to $y$. $\sigma_xy$ is the covariance of input image patches $x$ and $y$.
In the each comparison, it takes into account correlation coefficient $\frac{\sigma_{xy}}{\sigma_x\sigma_y}$, luminance distortion $\frac{2\bar{x}\bar{y}}{\bar{x}^2 + \bar{y}^2}$ and contrast distortion $\frac{2\sigma_x\sigma_y}{\sigma_x^2 + \sigma_y^2}$ between images given in Eqn.\ref{Q0}. It's range is between -1 and 1. 

\begin{equation} \label{Q0}
\begin{split}
    Q_0(x, y) 
     = \frac{\sigma_{xy}}{\sigma_x\sigma_y} * \frac{2\bar{x}\bar{y}}{\bar{x}^2 + \bar{y}^2} * 
    \frac{2\sigma_x\sigma_y}{\sigma_x^2 + \sigma_y^2}
\end{split}
\end{equation}

To calculate the fusion quality index, $Q_w$, the image is divided into square regions.  For all these regions, $Q_0$ scores are calculated. Also, some regions include more information than others.  Thus, to specify which region contributes the result more, local weights, $\lambda(w)$ are determined. In the \cite{qw},  $\lambda(w)$ is taken as $\frac{\sigma_x}{\sigma_x+\sigma_y}$. If the Eqn.\ref{Qw} takes edge responses of all images, then the result index becomes $Q_e$.

\begin{equation} \label{Qw}
\begin{split}
Q_w(a, b, f) = \frac{1}{W}\sum_{w \in W} \Bigg( \lambda(w) Q_0(a, f| w) + \\ (1 - \lambda(w))Q_0(b, f| w) \Bigg)
\end{split}
\end{equation}

\begin{equation} \label{loss}
\begin{split}
    L(a, b, f) = \alpha * (1 - Q_w(a, b, f)) + \beta * (1 - Q_e(a, b, f)) \\ +  \gamma * 0.5 * (MSE(a, f), MSE(b, f))
\end{split}
\end{equation}

The overall loss function is defined in the Eqn.\ref{loss}. $\alpha, \beta$, and $\gamma$ are the loss weights to normalize losses to make them equally important.

\subsection{Training Details} \label{details}
The entire network described in the previous sections is trained with Flir ADAS \cite{flir} and KAIST \cite{kaist} datasets which both of them include day and night conditions. In TNO, visible images are grayscaled whereas, in VIFB, most of the visible images are colored. In both datasets, there are 21 visible and infrared image pairs. In the training phase, Gaussian blur and Gaussian noise are added to the input images. Input image size is fixed to 256x256 and input images are normalized between -1 and 1. The initial learning rate is 0.001. ADAM optimizer is used with $\beta_1 = 0.9$, $\beta_2=0.99$ and $\epsilon = 10^{-8}$ as mentioned in the \cite{adam}. Also, a multi-step learning rate scheduler is used. Steps are decided as 10, 20, and 30, and the decay factor for the learning rate is 0.5. That is after reaching the above epoch numbers in the training, the learning rate is cut in half. 30 epoch is generally enough to train the network with stated data sets.  

All system is implemented in Pytorch 1.6.0 with CUDA 10.2. The computer has Intel i5 4460 CPU, Nvidia GTX 970 4GB GPU, and 8 GB system memory. Approximately, whole datasets include 20K samples and one epoch takes 30 mins with the above setup.


\begin{table}[h]
    \centering
    \caption{\textsc{Evaluation Metric Scores of 21 test images given in TNO\cite{TNO}}}
    \label{table:scores}
    \begin{tabular}{|c|c|c|c|}
        \hline
        Metrics & Qw  &  Qe  &  Qe*Qw    \\
        \hline
        DLF     & 0.67280  & 0.62082 & 0.41769 \\
        \hline
        DeepFuse & 0.72468   & 0.70971  & 0.51431 \\
        \hline
        Hybrid\_MSD & 0.68894 & 0.71466  & 0.49236 \\
        \hline
        Dual\_Branch & 0.68242 & 0.65022 & 0.44372 \\
        \hline
        Rfn-Nest & 0.66136 & 0.65125 & 0.43071 \\
        \hline
        Proposed & \textbf{0.75285} & \textbf{0.74463} &\textbf{0.56059} \\
        \hline
    \end{tabular}
\end{table}

\begin{figure}[t]
	\centering
	\includegraphics[width=0.5\textwidth, height=0.25\textheight]{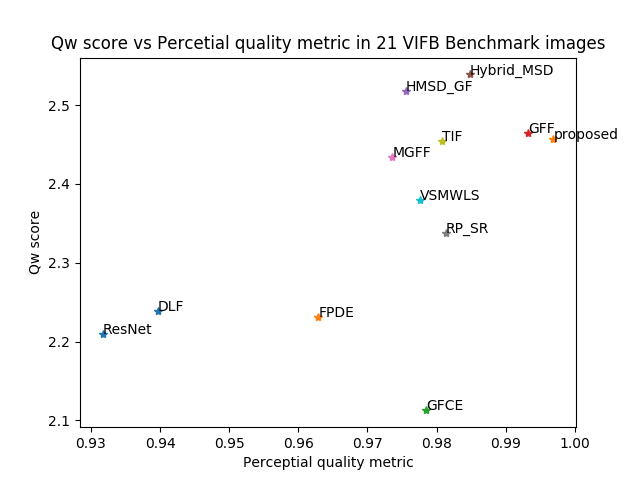}
	\caption{Quantitative comparison of several methods according to PaQ-2-PiQ\cite{PaQ} scores and Qw\cite{qw} scores. Bigger values mean better results for both metrics. 21 test image pairs are used given in TNO\cite{TNO}. }
	\label{fig:graph}
\end{figure}

\begin{figure}[t]
	\centering
	\includegraphics[width=0.45\textwidth, height=0.30\textheight]{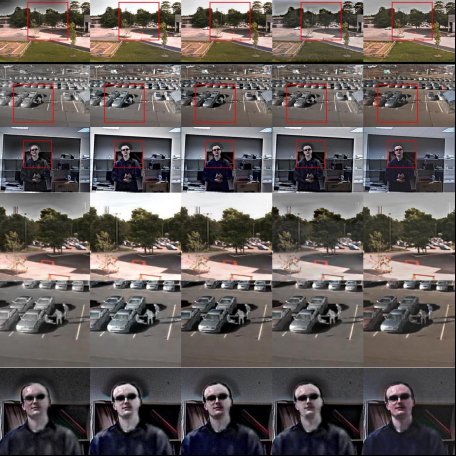}
	\caption{Qualitative comparison of several methods. Methods names are from left to right GFF \cite{gff}, HMSD\_GF \cite{hmsd-gf}, Hybrid\_MSD \cite{Hybrid-MSD}, TIF \cite{tif}, and proposed respectively. Image sequence names for the first 3 rows are carWhite, fight and labMan. The remaining rows are zoomed areas labeled in red rectangles in each sequence. }
	\label{fig:color}
\end{figure}

\begin{figure}[t]
	\centering
	\includegraphics[width=0.48\textwidth, height=0.30\textheight]{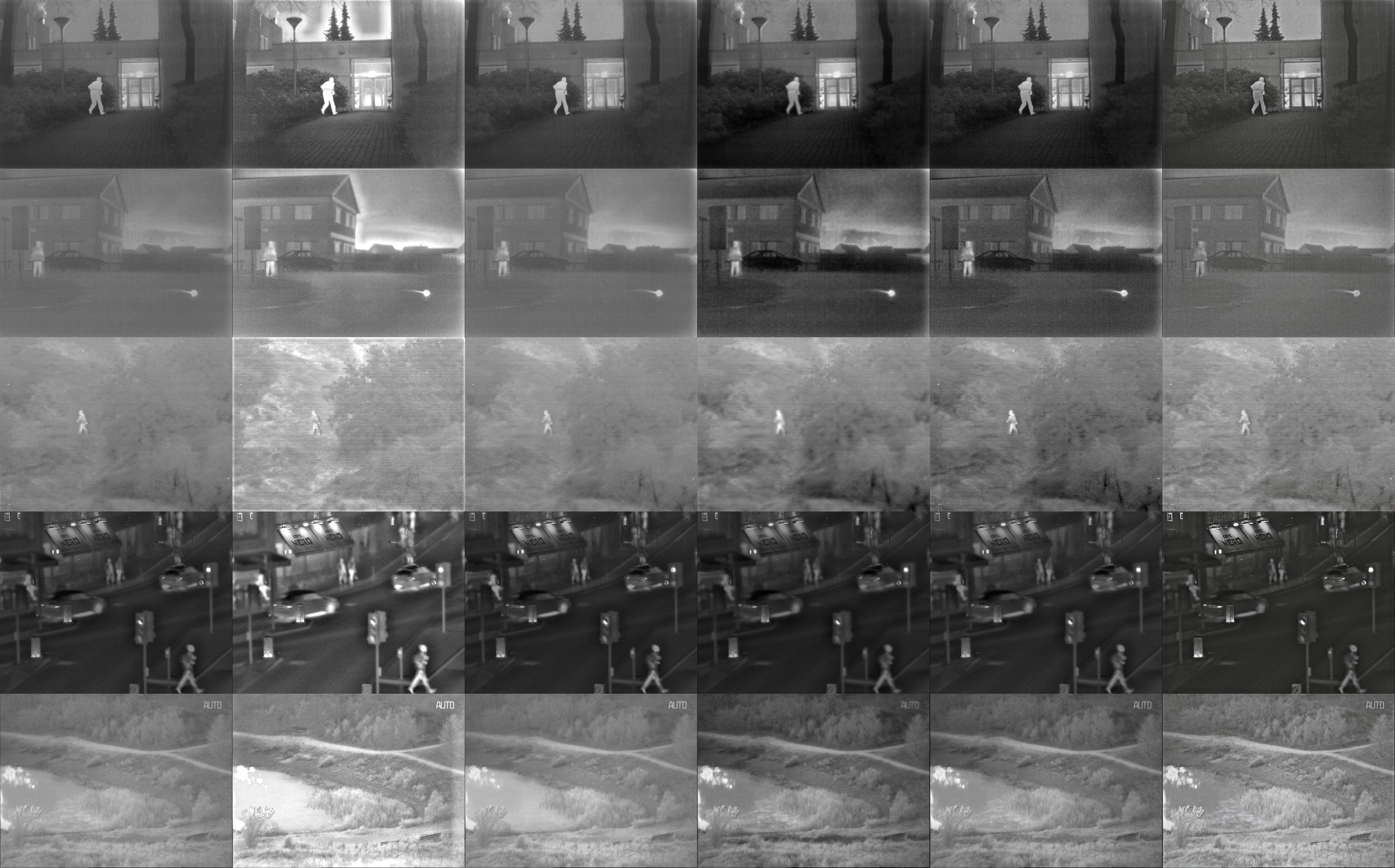}
	\caption{Qualitative comparison of several methods. Methods names are from left to right DLF \cite{DLF}, Hybrid\_MSD \cite{Hybrid-MSD}, Dual\_Branch \cite{dualbranch}, Rfn-Nest \cite{rfnnest}, DeepFuse \cite{deepfuse} and proposed respectively. Image sequences are taken from TNO dataset \cite{TNO}.}
	\label{fig:TNO}
\end{figure}

\section{Experiments}
Conducted experiment results are presented in this section. In the following sections, quantitative and qualitative comparisons are explained using TNO\cite{TNO} and VIFB\cite{vifb} datasets.

\subsection{Quantitative Performance Comparisons}
In this part, two comparison metrics are used: perceptual quality metric PaQ-2-PiQ\cite{PaQ}, and Qw\cite{qw} that is mentioned in the previous sections. PaQ-2-PiQ uses a neural network to predict the perceptual quality of the image. It is no reference method that means the score is predicted using a single input image only. In Fig.\ref{fig:graph}, our proposed method has the best perceptual quality score whereas it becomes the 4th method in the Qw score. However, as stated in VIFB\cite{vifb}, even if some methods have higher scores in evaluation metrics, there exist some artifacts or details that are not preserved well. Thus, in addition to fusion quality metrics, we added a perceptual quality metric to show the performance of our method more accurately. Furthermore, we conducted an experiment on TNO\cite{TNO} dataset using DLF \cite{DLF}, DeepFuse \cite{deepfuse}, Dual\_Branch \cite{dualbranch} and Rfn-Nest \cite{rfnnest} state-of-the-art deep learning methods and Hybrid\_MSD \cite{Hybrid-MSD} method. In Table \ref{table:scores}, average scores of the six methods, including the proposed method, are given and it is seen that our proposed method obtain the best results. For VIFB comparisons, images provided on their result page are used. For TNO comparisons, Hybrid\_MSD results are generated using provided MATLAB code and other results obtained from their project pages. 

Computational complexity and running time of a method are important when the application needs real-time performance. The run-time performance of our proposed method is better than any method mentioned above. The average inference time for both VIFB and TNO datasets is 0.0056 seconds. This is approximately 178 FPS in our setup given in Section \ref{details}. In VIFB and \cite{reviewDL}, the speed performance of numerous methods are compared. Among these results, the fusion method TIF takes 0.13 seconds which is the best result in VIFB. ResNet\cite{resnet} shows the best runtime performance which is 4.80 seconds as a deep learning based approach in VIFB. In \cite{reviewDL}, FLGC-fusionGAN\cite{flgc} shows best performance which is 0.0307 seconds. As seen, the proposed method has the lowest runtime(nearly x6 times faster) and it can be increased using some device-specific optimizations.

\subsection{Qualitative Performance Comparisons}
In Fig.\ref{fig:color}, VIFB dataset is used for comparison. To understand details better, red rectangle areas of each image is zoomed to compare easily. In the 4th row, the sky of the image is better fused in our method. HMSD\_GF and Hybrid\_MSD don't fuse images properly. The proposed method provides a more natural image with providing details. In the 5th and 6th rows, other methods do not construct colors well. This makes these images unrealistic and color information is lost. Thus, preserving important details without losing perceptual quality. Similar thoughts can be considered for Fig.\ref{fig:TNO}. Even if visible spectrum images are grayscaled for TNO dataset, proposed methods present more detailed and natural images.


\section{Conclusion}
We present a novel method to solve the visible and infrared spectrum image fusion problem. Our proposed methods are compared to the state-of-the-art methods both quantitatively and qualitatively. The proposed method is end-to-end trainable and there is no required additional processing.
In addition, we propose a new loss function that learns desired features when reconstructing the result. Also, our architecture is the fastest compared to other methods and can be used in embedded computer vision tasks such as object detection, segmentation, or tracking. As future work, the proposed method will be applied in Nvidia Jetson boards. Also, a more general dataset that includes different kinds of objects can be created to represent image spectrum information better.

\bibliographystyle{IEEEbib}
\bibliography{refs}

\end{document}